\documentclass[review]{elsarticle}	

\usepackage{slashbox}
\usepackage{setspace}
\usepackage{amsmath}
\usepackage{amssymb}
\usepackage{wasysym}
\usepackage{multirow}
\usepackage{subcaption}
\usepackage{algorithm}
\usepackage{algorithmic}
\usepackage{hhline}
\usepackage{adjustbox}
\usepackage{hyperref}
\usepackage{array}
\usepackage{amsthm}

\newtheorem{theorem}{Theorem}

\newtheorem{conjecture}{Conjecture}
\newtheorem{lemma}{Lemma}
\graphicspath{{figs/}}

\usepackage{tikz}

\journal{Neural Networks}

\begin{document}
\begin{frontmatter}
\title{Adaptive Low-Rank Regularization with Damping Sequences to Restrict Lazy Weights in Deep Networks}

\author{Mohammad Mahdi Bejani}
\address{Department of Mathematics and Computer Science, Amirkabir University of Technology (Tehran Polytechnic), Iran, e-mails: mbejani@aut.ac.ir}

\author{Mehdi Ghatee}
\address{Department of Mathematics and Computer Science, Amirkabir University of Technology (Tehran Polytechnic), Iran, e-mails: ghatee@aut.ac.ir}




\begin{abstract}
Overfitting is one of the critical problems in deep neural networks. Many regularization schemes try to prevent overfitting blindly. However, they decrease the convergence speed of training algorithms. Adaptive regularization schemes can solve overfitting more intelligently. They usually do not affect the entire network weights. This paper detects a subset of the weighting layers that cause overfitting. The overfitting recognizes by matrix and tensor condition numbers. An adaptive regularization scheme entitled Adaptive Low-Rank (ALR) is proposed that converges a subset of the weighting layers to their Low-Rank Factorization (LRF). It happens by minimizing a new Tikhonov-based loss function. ALR also encourages lazy weights to contribute to the regularization when epochs grow up. It uses a damping sequence to increment layer selection likelihood in the last generations. Thus before falling the training accuracy, ALR reduces the lazy weights and regularizes the network substantially. The experimental results show that ALR regularizes the deep networks well with high training speed and low resource usage.
\end{abstract}

\begin{keyword}
  Deep Neural Network, Regularization, Overfitting, Matrix Decomposition, Damping Coefficients, Lazy Weights.
\end{keyword}

\end{frontmatter}

\section{Introduction}
%
%
%
%
In the supervised machine learning, the complexity of the learning function $f$ is a critical concept. It can be defined as \cite{dem2015reg}:\begin{equation}\label{eq:tikhonov-regularization}
R(f) = \int \|\frac{\partial f}{\partial x} \|_2^2dx
\end{equation}
When the model complexity is high or $R(f)$ is significant, a small noise in the input causes a substantial change in the output, and the generalization fails, or overfitting occurs. In deep neural networks, because of their intrinsic complexity, the model tends to memorize the samples, and the generalization power reduces \cite{cawley2007preventing}. Different regularization methods are defined to impose a dynamic noise to the model through the training procedure \cite{bejanireview}. One of the most popular techniques is dropout \cite{srivastava2014dropout}, and its family \cite{wan2013regularization,kang2018shakeout,khan2018bridgeout,krueger2016zoneout,larsson2016fractalnet,khan2019regularization}. In these methods, in each iteration, a subset of weights of the neural network is selected to train. Some of these methods, such as \cite{kang2018shakeout,khan2018bridgeout}, impose small noise on the model weights. Using noise on the weights does not allow just memorizing the training data. Some other methods select a subset of loss functions to regularize the network \cite{bejani2021Laliga}.\\
However, many regularization techniques impose the noise on all components of the learning model blindly. They do not note the time and place of the overfitting occurrence. It is the reason for the slow convergence of their training process. Abpeikar et al. \cite{abpeikar2020adaptive} proposed an expert node in the neural trees to evaluate the overfitting, and when it was high, they used regularization. Bejani and Ghatee \cite{bejani2019convolutional} introduced the adaptive regularization schemes, including the adaptive dropout and adaptive weight decay to control overfitting in the deep networks. But their methods add some hyper-parameters to the learning procedure, and determining these hyper-parameters is not easy. Therefore, to determine the hyper-parameters concerning the overfitting, they used matrix decomposition to regularize the learning model \cite{bejani2020NNGhatee}. Various matrix decomposition methods such as spectral decomposition, nonnegative matrix factorization, and Low-Rank Factorization (LRF) have been proposed to summarize the data of a matrix (or a tensor) \cite{symeonidis2016matrix}. To apply these methods in the data mining fields, one can note \cite{ng2002spectral}. They are also good options for simplifying matrix weights in deep neural networks. In this regard, some attempts are very effective \cite{denton2014exploiting,jaderberg2014speeding,lebedev2014speeding}. In a recent paper, Tai et al. \cite{tai2015convolutional} used low-rank tensor decomposition to remove the redundancy in CNN kernels. Also, Bejani and Ghatee \cite{bejani2020NNGhatee} derived a theory to regularize deep networks dynamically by using Singular Value Decomposition (SVD). In adaptive version of this technique, namely ASR, the following loss function was minimized:
\begin{equation}\label{eq:ASRLOSS}
E^*(\theta_{t}) =E(\theta_{t})+ \gamma ||W - W_V^*||_F^2
\end{equation}
where $\theta_t=\{W_t, b_t\}$ show the synaptic weights and the bias vector, $\gamma$ is a regularization parameter, and $W_V^*$ is the SVD decomposition of the best-found weights on validation data. As mentioned by \cite{bejani2020NNGhatee}, ASR solves the overfitting substantially.\\
In continuation of these works, in this paper, we define a new measure based on the condition number of the matrices to detect when the overfitting occurs. We also identify the layers of a deep neural network that cause the overfitting problem. To solve this problem, we decompose the ill-posed matrices into low-rank matrices. We developed two new regularization schemes, namely Direct Low-Rank (DLR) and Adaptive Low-Rank (ALR). These methods can compete with the famous regularization methods in dropout family. These results will be supported by some experiments on small-size and large-scale datasets, separately. We also visualize the behavior of DLR on a noisy dataset to show how LRF solves overfitting. Finally, the results of ALR are compared with some famous regularization schemes including dropout methods \cite{kang2018shakeout,khan2018bridgeout,krueger2016zoneout,larsson2016fractalnet,khan2019regularization}, adaptive dropout method \cite{bejani2019convolutional}, weight decay with and without augmentation methods \cite{engstrom2017rotation,kwasigroch2017deep,wkasowicz2017computed,galdran2017data}.\\
In what follows, we present some preliminaries in Section 2. In sections 3 and 4, DLR and ALR are expressed. In Section 5, we present the results of empirical studies. The final section ends the paper with a brief conclusion.

\section{Preliminaries}
The overfitting of a supervised learning model such as a neural network is related to the condition number of the following nonlinear system:
\begin{equation}\label{eq:learning_label}
\sum_{i=1}^T\|f(x_i,\{W^1,...,W^L\}) - y_i\|_F^2 = 0,
\end{equation}
where $f$ is the output of the neural network with $L$ layers. $W^l$ is the weight matrix (or tensor) of the layer $l.$ $T$ is the number of training samples, and $(x_i,y_i)$ is the pair of the input and output of the $i^{th}$ sample. After solving this nonlinear system and finding $w_i$s, the learning model predicts the output for the unseen data. In numerical algebra, the condition number of a system shows the stability of the solution directly \cite{datta2010numerical}. When the condition number is great (very greater than 1), the system's sensitivity over the noise is very high, so the learning model's generalization ability decreases significantly. Thus, it is a good idea to evaluate the learning model's complexity by condition number \cite{datta2010numerical}. The condition number is defined for linear and nonlinear systems. For a linear system $Ax=b,$ where $A\in \mathbb{R}^{m\times n}$, $x\in \mathbb{R}^{n}$, and $b\in \mathbb{R}^{m}$, the condition number is defined as $\kappa(A) =\|A\|\|A^{-1}\|$ \cite{datta2010numerical}. Also, for the non-linear system $f(x) = y$ where $f$ is a non-linear vectorized function, the condition number is \cite{trefethen1997numerical}:
\begin{equation}\label{eq:non-linear-condition-number}
\kappa(f(\theta)) = \frac{\|J(\theta)\|_F \|\theta\|_F}{\|f(\theta)\|_F},
\end{equation}
where $\|.\|_F$  is Frobenius norm, $\theta$ is parameters of $f$, and $J(\theta)$ is Jacobian matrix of $f$ concerning $\theta$.

\subsection{Matrix factorization}
In this part, we discuss popular matrix factorization (decomposition) and show their ability to improve system stability. Consider an arbitrary matrix $A$ that is factorized into $r$ matrices $B_i$ and $A = \prod_{i=1}^r B_i$. In some instances, LU decomposition, Cholesky decomposition, Singular Value Decomposition (SVD), nonnegative matrix decomposition, and binary decomposition can determine the factors \cite{elden2019matrix}. Now, we focus on the Low-Rank Factorization (LRF) that approximates any matrix $A$ with two lower rank matrices $V$ and $U$. For a better approximation, the following optimization problem can be solved:
\begin{equation}\label{eq:LRF}
\min_{W,H} \|A - VU\|_F
\end{equation}
Thus, when $A\in R^{n,m},$ LRF factorizes $A$ into two matrices $V\in R^{n,k}$ and $U\in R^{k,m}$ and $Rank(U)=Rank(V)=k<Rank(A).$ To find $U$ and $V$ by optimization model \ref{eq:LRF}, see \cite{nimfa2020fact}.

\subsection{Tensor factorization}
There are two main approaches to factorize a tensor; explicit and implicit. In the explicit factorization of any tensor $T$, we try to find $r$ sets of vectors $a_i$, $b_i$ and $c_i$ such that $\sum_{i=1}^r (a_i b_i^T) \odot c_i$ approximates $T$, where $\odot$ is the tensor production. By minimizing $\|T - \sum_{i=1}^r (a_i b_i^T) \odot c_i\|_2^2,$ we can factorize $T$ into $r$ components \cite{tai2015convolutional}. However, the explicit tensor decomposition is an NP-hard problem\cite{hillar2013most}. Therefore, this type of decomposition is not the best way for the regularization of deep networks. Instead, in implicit factorization, we try to apply the matrix factorization methods directly. To this aim, any tensor $T$ is sliced into some matrices, and on every matrix, we apply the matrix factorization. Our results show the efficiency of this approach for deep learning regularization.
\subsection{Visualization of factorization effects}
\begin{figure*}
\begin{subfigure}[t]{0.32\linewidth}
\centering
\includegraphics[scale=0.37]{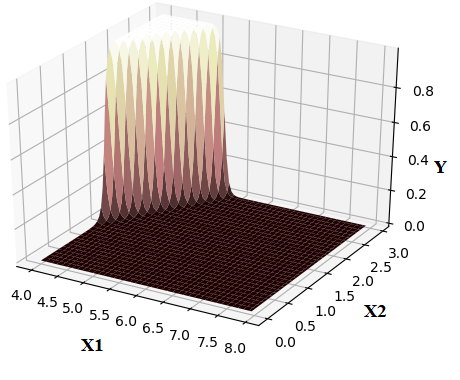}
\caption{Trained surface on original data.}
\label{fig:effect-of-lrf-1}
\end{subfigure}
\begin{subfigure}[t]{0.32\linewidth}
\centering
\includegraphics[scale=0.37]{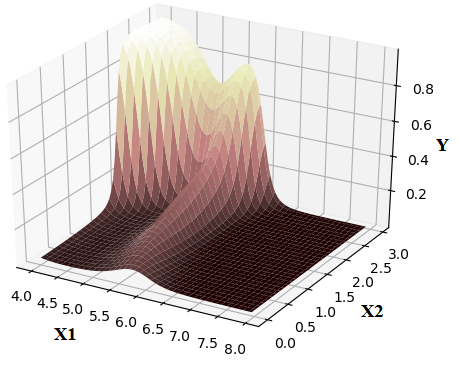}
\caption{Trained surface on noisy data.}
\label{fig:effect-of-lrf-2}
\end{subfigure}
\begin{subfigure}[t]{0.32\linewidth}
\centering
\includegraphics[scale=0.37]{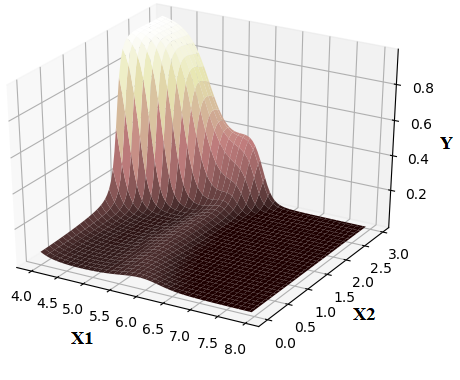}
\caption{Trained surface on noisy data when LRF regularization is used.}
\label{fig:effect-of-lrf-3}
\end{subfigure}
\caption{The learning model's surface that is trained by a Perceptron Neural Network (MLP) on Iris dataset.}
\label{fig:effect-of-lrf}
\end{figure*}

To visualize the effect of matrix and tensor factorization as the regularization method, we designed a test to show how they can improve the learning models. To this end, we used
a perceptron neural network with three hidden layers to train Iris dataset \cite{anderson1936species} and Iris data with high-level artificial noise. Figures \ref{fig:effect-of-lrf-1} and \ref{fig:effect-of-lrf-2} show two surfaces trained by the original dataset and noisy dataset, respectively. As one can see, the second learning model is over-fitted because of noisy data. In the third experiment, we substitute LRF of the weighting matrices instead of the second learning model's matrices for regularization purposes. Fig. \ref{fig:effect-of-lrf-3} shows the new surface. It is trivial that the regularized network is more similar to the original learning model without overfitting noisy data. Also, the model is simpler.

\begin{figure*}
\centering
\includegraphics[scale=0.35]{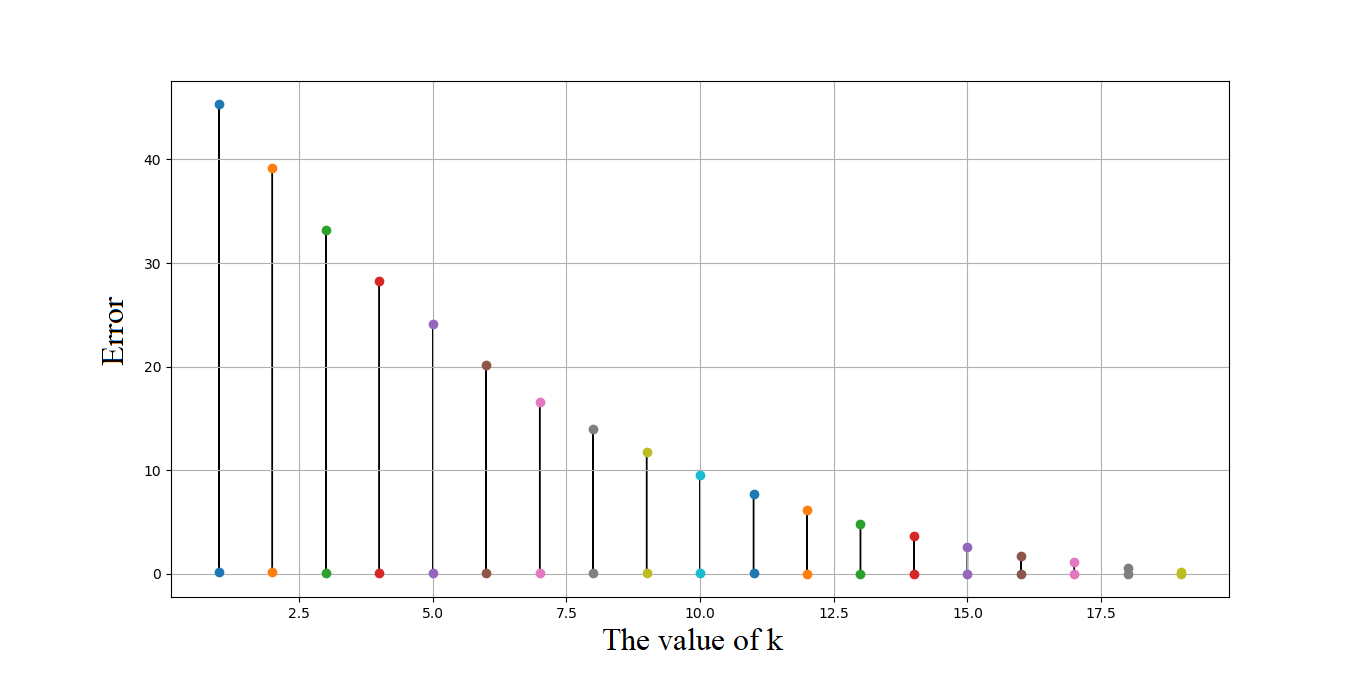}
\caption{Visualization of Theorem 1: For each $k$, the top point shows the error value after exchanging trained weights $W^*$ with $LRF_k(W^*)$, and the low point shows the error value after one retraining epoch.}
\label{fig:effect-th-1}
\end{figure*}
\section{Details of DLR}
DLR is a regularization technique that contains the following fundamental steps:
\begin{enumerate}
\item Detecting the overfitting in successive steps,
\item Identifying the matrices with a significant effect on overfitting and defining a probability distribution on these matrices,
\item Selecting some over-fitted matrices randomly based on the distribution achieved from the previous step,
\item Using LRF to regularize the selected matrices and the tensors.
\end{enumerate}
The DLR details are presented in Algorithm 1. This algorithm evaluates the overfitting dynamically \cite{bejani2019convolutional}. When the overfitting is small, the learning procedure continues; else, the regularization method affects the over-fitted model. Such a scheme saves the training speed and increases the generalization ability. The following measures the level of model overfitting dynamically:
\begin{equation}\label{eq:over-fitting-strict-computation}
v(t) = \frac{Error_{Validation}(t)}{Error_{Train}(t)},
\end{equation}
where $t$ is the iteration number. It is worth noting that $v(t)$ has an oscillatory behavior and iteratively decreases and increases. Therefore, we get the average of the last three $v(t)$. Instead of three, we can use a hyper-parameter $p$ as \textit{patient of regularization}. When the overfitting is recognized, the learning algorithm identifies the overfitting reason and tries to resolve this problem. Because of the layered architecture of deep networks, it is possible to find some of the layers that cause overfitting. We regularize them by LRF to remove the effects of data noises in the weighting matrices. However, it keeps the major trend of data. It leads to a softer surface (as mentioned in Fig.\ref{fig:effect-of-lrf-3}).

Finding a sub-set of the layers with the highest effect on overfitting is hard. Instead, we return to the training system \ref{eq:learning_label} and compute the complexity of each layer by its condition number \ref{eq:non-linear-condition-number}. Denote the condition number of $l^{th}$ layer with $\kappa(f_l)$. We are ready to regularize the weighting matrices with great $\kappa(f_l).$ But, the experiments show that regularization on each over-fitted matrix increases the processing time. Thus, similar to dropout \cite{srivastava2014dropout}, we sample a random number by $Bernoulli(\Gamma(f_l))$ distribution, that is defined by Eg. (\ref{eq:normalize_the_cd}). When the produced random number is less than the following normalized parameter, we use LRF regularization to simplify the weighting matrices:

\begin{equation}\label{eq:normalize_the_cd}
\Gamma(f_l) = \frac{\kappa(f_l)}{\max_i \kappa(f_i)}.
\end{equation}
Because the convolution layers are tensors, they are sliced to matrices with $FS \times FN$, where $FS$ and $FN$ are the filter size and the filter number, respectively. Then, the algorithm approximates these matrices directly.

Furthermore, LRF can approximate any matrix with its $k-$rank factorization. The following theorem shows that LRF regularization is independent of the value of $k$ in linear learning models.
\begin{figure*}
\centering
\includegraphics[scale=0.49]{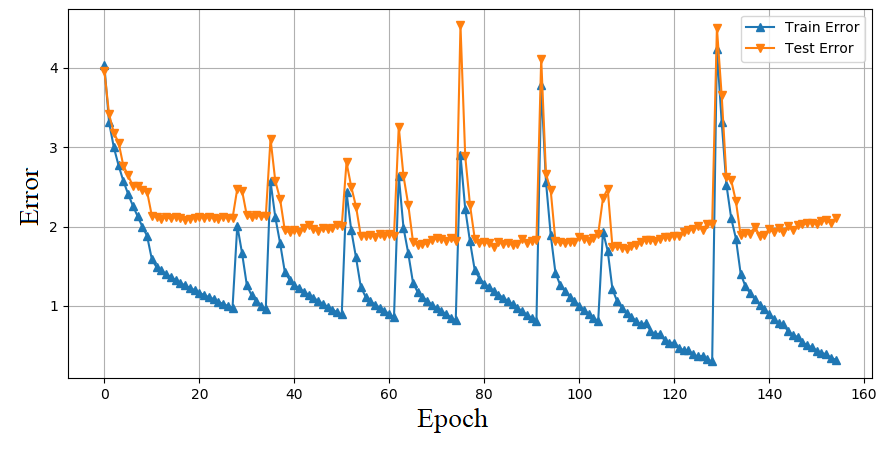}
\caption{The training and testing errors of a non-linear network by getting LRF in some epochs of Algorithm 1.}
\label{fig:lrf_th_nonlin}
\end{figure*}
\begin{theorem}\label{th:lrf}
Assume $f(x;W) = Wx+b$ is a neural network with linear activation function, $W$ is the weights matrix, and $x$ is an input vector with zero and $\sigma^2$ as the mean and the variance. Assume all features of $x$ are independent. Let $W^*$ shows the optimal training weights. For any $k < rank(W^*)$, if $W_k = LRF_k(W^*)$ is substituted instead of $W^*$ in the network, after one epoch of the retraining, the optimal weights converge to $W^*$.
\end{theorem}
\textit{Proof.} See appendix \ref{app:th1}\\
For visualizing the effect of \textbf{Theorem 1}, we trained a linear network on a dataset with $\mathcal{N}(\vec{0},I)$ distribution and found the optimal weights $W^*$. Now, for any $k \in \{1,...,20\}$, we defined $W_k=LRF_k(W^*)$ and substitute $W_k$ instead of $W^*$. Then, we retrained the network one epoch and found the loss values again. Fig. \ref{fig:effect-th-1} shows that, for all $k$, the loss values converge to the originally trained network's loss value. 
\\ Besides, some evidence shows the prediction of Theorem 1 meets for non-linear cases after several retraining epochs. For instance, we trained a network on the SVHN dataset by Algorithm 1 and got LRF from some weights along the training epochs. Fig.\ref{fig:lrf_th_nonlin} shows the network's loss values on the training and the testing datasets. After substituting LRF instead of the weights, the network's loss values increase, but after a few epochs, they converged to the error before LRF getting. But the returning does not happen in one epoch, similar to the linear models. This evidence supports the following conjecture.
\begin{conjecture}\label{th:non-linear-effect}
Assume $f(x;W)$ is a neural network with non-linear activation function and $x \sim \mathcal(\vec{0}, \Sigma)$, where $\vec{0}$ is a zero vector as the means, and $\Sigma$ is the covariance of features of the training dataset. Suppose that $f(x;W)$ is trained and the optimal weights are $W^*$ with the optimal error $E(W^*)$. Assume $W^*$ is exchanged by $W_k = LRF(W^*)$ and the network is retrained on the dataset with $O(L \|\Sigma\|_2)$ epochs. Then, the network weights $W^{**}$ and the network error $E(W^{**})$ satisfy the following:
\[E(W^{**}) \leq E(W^*)\]. 
\end{conjecture}
The preferences of DLR regularization compared with ASR \cite{bejani2020NNGhatee} can be summarized as the following:
\begin{enumerate}
\item It is independent of the validation dataset, and DLR detects the overfitting by condition number.
\item 
In each training epoch, some weight matrices (in our experiments, usually 25\% of the weights) are substituted with their LRF. This selection increases the training speed and decreases resource usage.
\item 
Based on Theorem 1, we set $k=1$ in all experiments. Thus the approximation of matrices can be computed by two vectors, and the computations become faster than ASR.
\item The vanishing of SVD \cite[Theorem 4]{bejani2020NNGhatee} does not happen in DLR because the weights are randomly selected based on a Bernoulli distribution concerning the condition number.
\end{enumerate}
\begin{algorithm}
\caption{Training Algorithm with DLR}
\begin{algorithmic}[1]
\REQUIRE $\alpha_t$: Step-size
\REQUIRE $D_t$: Improvement Direction
\REQUIRE $t \leftarrow 0$: Time-step
\REQUIRE $\theta_t \leftarrow \{W_t,b_t\}$: Weights and biases of the neural network.
\REQUIRE $E(\theta_t)$: Error function of the network.
\REQUIRE $W$: number of the weights in the network.
\REQUIRE $L$: number of the trainable layers in the network.
\REQUIRE  $\epsilon$: an small positive threshold.

\WHILE{$\|\bigtriangledown_\theta E(\theta_t) \|>\epsilon$}
	\STATE $g_{t+1}\leftarrow \bigtriangledown_\theta E(\theta_t)$
	\STATE $D_{t+1} \leftarrow DescentDirection(g_{t+1}, \alpha_{t+1})$ that is a function that returns an improvement direction based on the inputed gradient and step-size $\alpha_{t+1}$.
	\STATE $\theta_{t+1} \leftarrow \theta_t -D_{t+1}$ 
	\STATE $v(t) \leftarrow \dfrac{E_{Validation}(t)}{E_{Train}(t)}.$ 
	\IF{$v(t)$ increases}
		\FOR{ \text{all layers} $l \in \{1,...,L \}$ and any weights tensor $W^l$ and bias vector $b^l$} 
			\STATE Compute the $\kappa(f_l)$ by Eq. \ref{eq:non-linear-condition-number}.
		\ENDFOR
		\FOR{ \text{all layers} $l \in \{1,...,L \}$ and any weights tensor $W_{l}$} 
			\STATE Compute $\Gamma(f_l)$ based on Eq. \ref{eq:normalize_the_cd}
			\STATE $r \leftarrow rand(0,1)$
			\IF{$r \leq \Gamma(f_l)$}
				\STATE $W^l \leftarrow LRF(W^l)$.
			\ENDIF
		\ENDFOR
		\STATE $b^l_{t+1} \leftarrow b^l_{t}$
	\ENDIF
	\STATE $t \leftarrow t + 1$
\ENDWHILE
\RETURN{$\theta_{t}$}
\end{algorithmic}
\end{algorithm}

\section{Details of ALR}
The experimental results show that DLR prefers the other regularizers for shallow networks and can compete with other regularization schemes. However, DLR results in the deep networks are not the best when the model complexity is high. To solve this problem, the DLR is combined with the Thikonov regularization scheme presented in Eq. (\ref{eq:ASRLOSS}). To this end, we minimize the following new loss function:
\begin{equation}\label{eq:lossthikhonov}
E^*(\theta_t) = E(\theta_{t-1}) +  \gamma \sum_{l\in \mathcal{W}} \|W_{t-1}^l - LRF(W_{t-1}^l)\|_F^2,
\end{equation}
where $\mathcal{W} \subseteq \{1,...,L\}$ is a sub-set of the layers, $W^l$ is the weights of the $l^{th}$ layer, $\theta_t$ is the network weights in epoch $t$, and $\gamma = \frac{1}{|\mathcal{W}|}$ is a fixed parameter. In each training epoch, a new $\mathcal{W}$ is selected based on $Bernoulli(\Gamma(f^l))_{l\in\{1,...,L\}}$. If the condition number of $l^{th}$ layer is great, its chance for participating in the regularization term increases.

ALR method found and solved the following three problems:
\begin{itemize}
\item \cite{bejani2020NNGhatee} proved that the effect of matrix approximation in ASR decreases when training epochs grow up. Then, the regularization term's effect is scarce at the end of the training procedure, while the overfitting problem possibly occurs. The same problem remains in the DLR method.
\item The matrix approximation in the last training epochs corrupts the learning model's performance. 
\item The third problem happens because of lazy weights. It is worth noting that some of the network weights do not contribute to the regularization process along the training epochs. Probably, they appear at the last training epochs. However, the training terminates soon, and they have not enough occasion to tune-up. We entitled them as lazy weights. We increase their selection probability when the epochs grow. This approach encourages them to contribute in more regularization steps. A damping sequence $DS(t)$ is used for epoch $t$ to increase the likelihood of layer selection in the last training epochs:
\begin{equation}\label{eq:probdamp}
\text{Prob($f_l$ selection at epoch t)}=\frac{\Gamma(f_l)}{DS(t)}.
\end{equation}
The resulted ALR does probably similar to ASR in more training epochs and restricts lazy weights occurrence. We experimentally show that $DS(t)=\frac{1}{\log{(t)}}$ is a suitable damping sequence when $t\geq 2$. In the following, we show the effect of lazy weights on training procedures.
\end{itemize}

Details of ALR is given in Algorithm 2. To analyze ALR, we study the following.
\begin{algorithm}
\caption{Training Algorithm with ALR}
\begin{algorithmic}[1]
\REQUIRE Damping sequence $DS(t)$.
\REQUIRE Similar to Algorithm 1.
\WHILE{ $\|\bigtriangledown_\theta E^*(\theta_t) \|>\epsilon$}
\STATE  $\mathcal{W} \leftarrow \emptyset$
	\FOR{ \text{all layers} $l \in \{1,...,L \}$ and any weights tensor $W^l$} 
		\STATE Compute $\Gamma(f_l)$ based on Eq. \ref{eq:normalize_the_cd}
		\STATE $r \leftarrow rand(0,1)$
		\IF{$DS(t)r \leq \Gamma(f_l)$}
			\STATE $\mathcal{W} \leftarrow \mathcal{W} \cup \{l\}$.
		\ENDIF
	\ENDFOR
	\STATE $g_{t+1}\leftarrow \bigtriangledown_\theta \Big[E(\theta_t) + \frac{1}{|\mathcal{W}|}\sum_{l\in \mathcal{W}}\|W^l_t - LRF(W^l_t)\|_F^2\Big]$
	\STATE Do similar to Steps 3 and 4 of Algorithm 1.
	\STATE $t \leftarrow t + 1$
\ENDWHILE
\RETURN{$\theta_{t}$}
\end{algorithmic}
\end{algorithm}

\begin{lemma}
If $D = \{x_i,y_i\}_{i=1}^T$ is an arbitrary classification dataset and $f(x;\tilde{W}) = y$ is a linear classifier with trained weights $\tilde{W} \in \mathbb{R}^{m\times n}$, then rank of $\tilde{W}$ is greater than one.
\end{lemma}
\textit{Proof.} See appendix \ref{app:lem1}

\begin{theorem}
If a neural network, $f(x;\theta)$, is trained on an arbitrary dataset with ARL, and $W^*$ is a lazy weight that is selected for the ARL regularization term, then the model accuracy drops in the next training iteration.
\end{theorem}
\textit{Proof.} See appendix \ref{app:th3}

Finally, by considering \cite[Theorem 1]{lu2021localdrop}, we can directly conclude that Rademacher complexity \cite{bartlett2001rademacher} of the neural networks decreases by using either DLR or ALR. Actually, in Step 14 of DLR, the LRF of matrices are substituted as the original weights. Also in Step 10 of ALR, the weights are implicitly converge to their LRF, because when $g_{t+1}\rightarrow 0$, both $E(\theta_t)$ and $\|W^l_t - LRF(W^l_t)\|_F^2$ converges to zero for all layer $l\in \mathcal{W}$. Thus, the results are direct.
\section{Empirical studies on DLR}
In this section, the DLR results are compared with the other regularization methods. Also, we check its performance to control the overfitting in the different datasets. DLR is applied on both shallow and deep networks. To use DLR implementation, one can refer to  \href{https://github.com/mmbejani/MatrixFactorizationRegularization}{Github}\footnote{https://github.com/mmbejani/MatrixFactorizationRegularization}.

\subsection{Effect of condition number in DLR}
To present the effect of condition number expressed in Eq. \ref{eq:normalize_the_cd} in the performance of DLR,  consider the following scenarios:

\begin{enumerate}
\item The first $k$ layers of the network are regularized when the overfitting occurs. In this scenario, the weight matrices of the first $d$ layers are factorized by LRF, and their approximations are substituted as the new weights matrices.
\item 
The last $d$ layers of the network are used for the same regularization plan.
\item This scenario is a hybrid of the random selection and standard DLR presented in Algorithm 1.
\end{enumerate}

VGG-19 network was trained on CIFAR-100 in these experiments. To trace these scenarios, we define the following criterion, namely summation of normalized condition number (SNCN):

\begin{equation}
SNCN(f) = \sum_{l=1}^L \Gamma(f_l),
\end{equation}

where $\Gamma(f_l)$ is defined in Eq.\ref{eq:normalize_the_cd} as the condition number of layer $l$. The smaller this criterion in different iterations, the greater the network's stability against overfitting.

\begin{figure}
\centering
\includegraphics[scale=0.4]{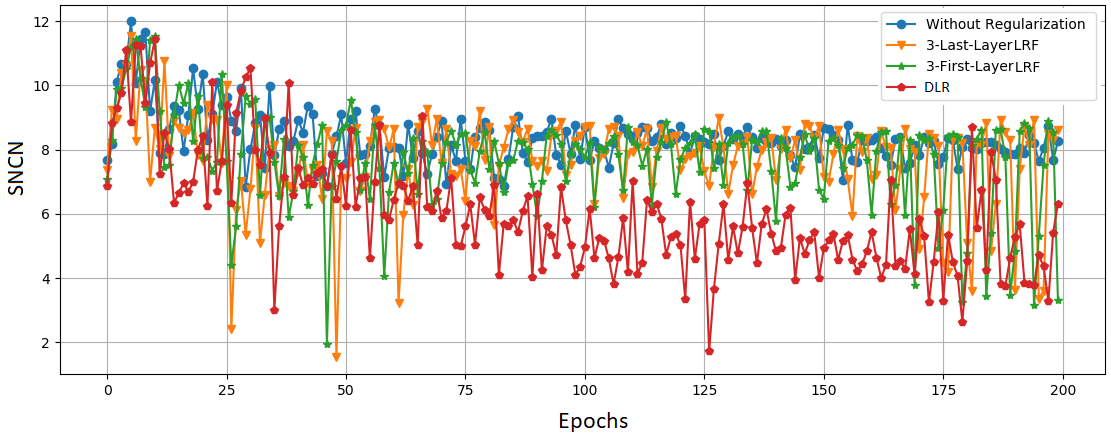}
\caption{SNCN criterion for each epoch of VGG-19 training on CIFAR-100  for three scenarios. }
\label{fig:scnarios_vgg_cd}
\end{figure}

\begin{figure}
\centering
\includegraphics[scale=0.33]{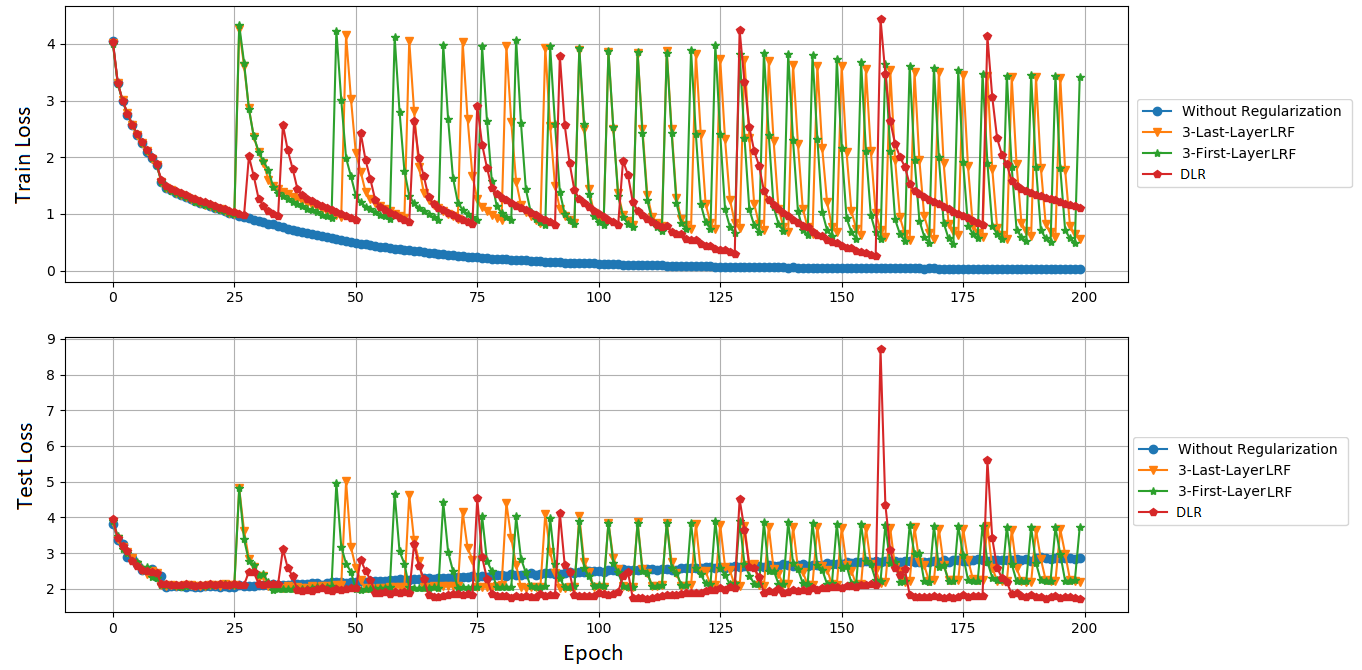}
\caption{The training and testing loss values of VGG-19 on CIFAR-100 in successive epochs for three scenarios.}
\label{fig:scnarios_vgg_per}
\end{figure}

In Figures \ref{fig:scnarios_vgg_cd} and \ref{fig:scnarios_vgg_per}, the performances of the VGG-19 for these scenarios are presented. As one can see, the third scenario archives the best results based on SNCN and testing loss values. However, the result of DLR on the training loss is worse than the others. As one can see, DLF in some epochs imposes a noise on training loss to prevent overfitting.

In another experiment,  a snapshot of the training results of DLR on Wide-Resnet network on CIFAR-10 is presented. It clears that the jumps of the loss functions in the training and the testing results are different, and DLR affected some parts of the network, where it is over-fitted. Thus, DLR simplifies some parts of the over-fitted model and seldom removes the useful information.

\subsection{Performance of DLR on Shallow Networks}
In this part, we show the results of DLR on the shallow networks applying different datasets. The used networks contain at most five layers, including several dense layers and one convolution layer. DLR results for training these networks are compared with training without regularization and training with dropout ($p=0.1,0.2,0.3$). Each experiment is repeated five times, and their average is presented in Table \ref{tab:small_dataset_comp}. In datasets Arcene, BCWD, and IMDB Reviews, DLR overcomes others. In the BCWD dataset, whose overfitting is low, DLR is somewhat weaker than training without regularization.
\begin{table}
\begin{center}
\caption{The comparison of the training with DLR and other regularization methods on shallow networks.}
\label{tab:small_dataset_comp}
\begin{tabular}{c|c|c|c|c|c}
\hhline{======}
Dataset Name & Regularization & Train A & Train L & Test A & Test L \\ 
\hhline{======}
\multirow{5}{*}{Arcene} & None & 99.4\%& $6.04\times 10^{-3}$ & 74.6\%& 0.202\\ \cline{2-6}
& Dropout (0.1) & 60.6\% & $3.87\times 10^{-1}$ & 59.2\% & 0.406\\ \cline{2-6}
& Dropout (0.2) & 60.0\% & $3.80\times 10^{-1}$ & 59.0\% & 0.411\\ \cline{2-6}
& Dropout (0.3) & 61.1\% & $3.72\times 10^{-1}$ & 58.0\% & 0.413\\ \cline{2-6}
& DLR & 97.4\% & $6.80\times 10^{-2}$ & \textbf{85.8\%} & \textbf{0.125}\\ \hline
\multirow{5}{*}{BCWD} & None & 98.7\% & $1.03\times 10^{-2}$ & 95.0\% & 0.038 \\ \cline{2-6}
& Dropout (0.1) & 99.8\% & $3.02\times 10^{-3}$ & 95.3\% & 0.037\\ \cline{2-6}
& Dropout (0.2) & 93.1\% & $4.50\times 10^{-2}$ & 90.4\% & 0.068\\ \cline{2-6}
& Dropout (0.3) & 90.8\% & $6.60\times 10^{-2}$ & 86.5\% & 0.112\\ \cline{2-6}
& DLR & 97.6\% & $1.86\times 10^{-2}$ & \textbf{95.6\%} & \textbf{0.032}\\ \hline
\multirow{5}{*}{BCWP} & None & 98.3\% & $1.22\times 10^{-2}$ & \textbf{76.4\%} & \textbf{0.205}\\ \cline{2-6}
& Dropout (0.1) & 97.1\% & $2.73\times 10^{-2}$ & 75.0\% & 0.229\\ \cline{2-6}
& Dropout (0.2) & 95.0\% & $3.52\times 10^{-2}$ & 71.9\% & 0.220\\ \cline{2-6}
& Dropout (0.3) & 88.0\% & $8.96\times 10^{-2}$ & 73.2\% & 0.211\\ \cline{2-6}
& DLR & 89.2\% & $8.63\times 10^{-2}$ & 74.6\% & 0.210\\ \hline

\multirow{5}{*}{IMDB Reviews} & None & 100.0\% & $7.60\times 10^{-4}$ & 85.3\% & 0.396\\ \cline{2-6}
& Dropout (0.1) & 99.7\% & $1.14\times 10^{-2}$ & 84.5\% & 0.542\\ \cline{2-6}
& Dropout (0.2) & 99.8\% & $1.06\times 10^{-2}$ & 83.9\% & 0.556\\ \cline{2-6}
& Dropout (0.3) & 99.7\% & $1.28\times 10^{-2}$ & 84.1\% & 0.538\\ \cline{2-6}
& DLR & 99.4\% & $3.54\times 10^{-2}$ & \textbf{85.6\%} & \textbf{0.354}\\ 
\hhline{======}
\end{tabular}
\end{center}
\end{table}

\subsection{Performance of DLR on deep networks}
In this part, VGG family is trained with DLR on CIFAR-10 \cite{krizhevsky2009learning} and CIFAR-100\cite{krizhevsky2009learning}.  Table \ref{tab:dlr-results} shows the results without  augmentation. Since the functionality of DLR is similar to dropout family schemes, we compare it with Dropout \cite{srivastava2014dropout}, Dropconnect  \cite{wan2013regularization}, Shakeout \cite{kang2018shakeout}, Bridgeout \cite{khan2018bridgeout}, and Spectral Dropout \cite{khan2019regularization} in this table. When the model complexity decreases, the DLR's performance is less than some of these regularization schemes. Therefore, DLR is useful when the model complexity is high.

\begin{table}
\begin{center}
\caption{The performance of the VGG family on CIFAR-10 and CIFAR-100 datasets using different regularization schemes (Boldfaces show the best accuracies)}
\label{tab:dlr-results}
\begin{tabular}{c|c|c|c|c|c|c}
\hhline{=======}
\backslashbox{Model}{Regularization} & DLR & \cite{srivastava2014dropout} & \cite{wan2013regularization} &  \cite{kang2018shakeout} & \cite{khan2018bridgeout} & \cite{khan2019regularization}
\\ \hhline{=======}
\multicolumn{7}{c}{CIFAR-10} \\ \hline
VGG-11 & \textbf{85.4\%} & 83.1\% & \textbf{85.4\%} & 85.2\% & 83.7\% & 85.2\% \\ \hline
VGG-13 & 84.8\% & 84.2\% & 84.2\% & \textbf{86.5\%} & 85.3\% & 85.7\% \\ \hline
VGG-16 & 85.1\% & 84.1\% & 85.7\% & 84.6\% & 86.3\% & \textbf{86.4\%} \\ \hline 
VGG-19 & 85.1\% & 83.3\% & 85.0\% & 84.6\% & 84.9\% & \textbf{85.4\%} \\ \hline
\multicolumn{7}{c}{CIFAR-100} \\ \hline
VGG-11 & 51.4\% & 52.3\% & \textbf{54.9\%} & 54.4\% & 53.9\% & 53.3\% \\ \hline
VGG-13 & 52.3\% & 52.6\% & 53.5\% & \textbf{55.1\%} & 54.7\% & 54.6\% \\ \hline
VGG-16 & 52.5\% & 53.4\% & 56.3\% & \textbf{55.5\%} & 54.4\% & 55.3\% \\ \hline 
VGG-19 & 52.2\% & 52.7\% & 54.5\% & \textbf{55.4\%} & 54.6\% & 55.9\% 
\\ \hhline{=======}
\end{tabular}
\end{center}
\end{table}

\section{Empirical studies on ALR}
Here, to show the preference of ALR  on the other regularizations, we compare their performance, resource usage, and convergence speed. In the first part, we experimentally show that $1/\log(t)$ is a suitable damping sequence to avoid lazy weights. In the second part, ALR is compared with other regularization schemes with this damping sequence. To see the implementation of ALR refer to \href{https://github.com/mmbejani/TikhonovRegularizationTerm}{Github}\footnote{https://github.com/mmbejani/TikhonovRegularizationTerm}.
\subsection{Damping Sequence}
Five scenarios for damping sequences are compared to find a suitable damping sequence for avoiding lazy weights. For each sequence, four networks on two different datasets are trained.  Table \ref{tab:damping-sequence} shows the experimental results. As one can see, sequence $1/\log(t)$ controls the lazy weights well, and in most of the scenarios, it improves the performance. Thus, we use this damping sequence for the next experiments. 

\begin{figure}
\begin{subfigure}[t]{1.0\linewidth}
\centering
\includegraphics[scale=0.34]{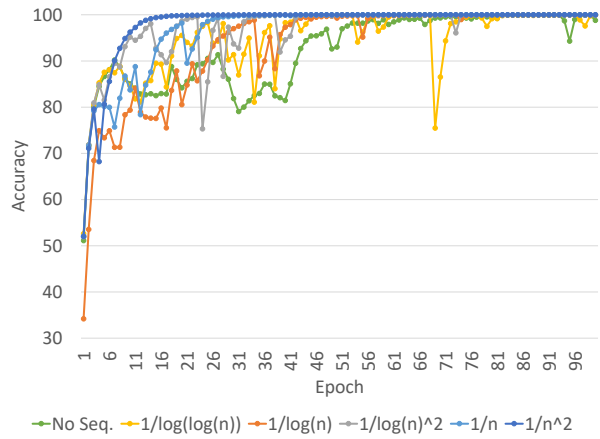}
\caption{Training Accuracy}
\label{fig:damping-train}
\end{subfigure}
\begin{subfigure}[t]{1.0\linewidth}
\centering
\includegraphics[scale=0.3]{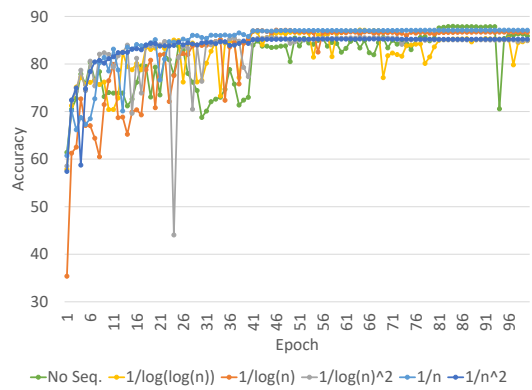}
\caption{Testing Accuracy}
\label{fig:damping-test}
\end{subfigure}
\caption{The performance of ALR with different damping sequences on the VGG-16 network on Caltech-101.}
\label{fig:damping}
\end{figure}

Besides, Fig. \ref{fig:damping} shows the effect of different damping sequences on the training and testing accuracies of VGG-16 on dataset Caltech-101 \cite{fei2004learning}.  Without any damping sequence, in $93^{th}$ epoch, some lazy weights are contributed, and the accuracy drops from $88\%$ to $71\%$. A weak damping sequence such as $\log(\log(t))$ behaves similarly. However, other damping sequences do not impose lazy weights, and the model accuracy has not dropped.

\begin{table}
\begin{center}
\caption{The VGG performances on CIFAR-10 and CIFAR-100 datasets using different damping sequences (Boldfaces show the best accuracies)}
\label{tab:damping-sequence}
\begin{tabular}{c|c|c|c|c|c}
\hhline{======}
\backslashbox{Model}{Regularization} & None & $1/\log(\log(t))$ & $1/\log(t)$ & $1/t$ & $1/t^2$
\\ \hhline{======}
\multicolumn{6}{c}{CIFAR-10} \\ \hline
VGG-11 & 83.5\% & 83.4\% & \textbf{86.7\%} & 86.5\% & 85.5\% \\ \hline
VGG-13 & 84.2\% & 84.5\% & \textbf{87.3\%} & \textbf{87.3\%} & 86.2\% \\ \hline
VGG-16 & 82.8\% & 84.2\% & \textbf{88.7\%} & 88.3\% & 86.3\% \\ \hline 
VGG-19 & 83.1\% & 85.3\% & 89.0\% & \textbf{89.1\%} & 86.7\%  \\ \hline
\multicolumn{6}{c}{CIFAR-100} \\ \hline
VGG-11 & 52.5\% & 52.4\% & 54.3\% & 54.2\% & \textbf{54.5\%} \\ \hline
VGG-13 & 52.1\% & 52.3\% & \textbf{55.4\%} & 54.3\% & 54.1\% \\ \hline
VGG-16 & 52.5\% & 52.8\% & \textbf{55.2\%} & 54.8\% & 54.3\% \\ \hline 
VGG-19 & 52.9\% & 52.5\% & 54.9\% & \textbf{55.0\%} & 54.7\%
\\ \hhline{======}
\end{tabular}
\end{center}
\end{table}
\begin{table*}
\begin{center}
\caption{The comparison between ALR and the different Tikhonov-based regularization schemes on the VGG family on different datasets (Boldfaces show the best accuracies)}
\label{tab:tikhonovcomparison}
\resizebox{1.0\textwidth}{!}{\begin{tabular}{c|c|c|c|c|c|c|c|c|c}
\hhline{==========}
\backslashbox{Model}{Regularization} & ALR & \cite{bejani2020NNGhatee} & \cite{krogh1992simple} & \cite{ayinde2019regularizing} & \cite{zou2005regularization} & \cite{phan2019group} & \cite{wu2014batch} & \cite{ma2019transformed} & None
\\ \hhline{==========}
\multicolumn{9}{c}{CIFAR-10} \\ \hline
VGG-11 & \textbf{86.7\%} & 85.7\% & 85.7\% & 85.6\% & 85.6\% & 84.3\% & 84.2\% & 84.9\% & 81.6\% \\ \hline
VGG-13 & \textbf{87.3\%} & 86.4\% & 84.1\% & 84.0\% & 85.3\% & 84.1\% & 83.7\% & 84.5\% & 82.1\% \\ \hline
VGG-16 & \textbf{88.7\%} & 86.4\% & 85.5\% & 84.6\% & 84.8\% & 86.0\% & 85.1\% & 84.0\% & 82.3\% \\ \hline
VGG-19 & \textbf{89.0\%} & 87.8\% & 83.6\% & 85.5\% & 83.9\% & 85.2\% & 86.4\% & 84.5\% & 82.3\% \\ \hline
\multicolumn{9}{c}{CIFAR-100} \\ \hline
VGG-11 & \textbf{54.3\%} & 53.0\% & 50.1\% & 51.5\% & 51.9\% & 51.7\% & 51.4\% & 50.5\% & 45.3\% \\ \hline
VGG-13 & \textbf{55.4\%} & 53.8\% & 52.1\% & 51.6\% & 52.1\% & 53.3\% & 51.4\% & 52.5\% & 47.3\% \\ \hline
VGG-16 & \textbf{55.2\%} & 54.6\% & 51.2\% & 51.9\% & 52.3\% & 53.1\% & 52.3\% & 52.3\% & 47.6\% \\ \hline
VGG-19 & \textbf{54.9\%} & 53.3\% & 52.5\% & 51.9\% & 52.8\% & 51.2\% & 52.1\% & 52.6\% & 48.9\% \\ \hline
\multicolumn{9}{c}{Caltech-101} \\ \hline
VGG-11 & \textbf{65.6\%} & 64.6\% & 62.5\% & 64.7\% & 65.3\% & 63.2\% & 64.0\% & 64.3\% & 60.5\% \\ \hline
VGG-13 & \textbf{65.9\%} & 65.1\% & 63.7\% & 62.9\% & 63.1\% & 64.2\% & 63.2\% & 64.4\% & 61.0\% \\ \hline
VGG-16 & \textbf{67.0\%} & 65.3\% & 65.2\% & 64.9\% & 63.3\% & 62.6\% & 65.1\% & 64.0\% & 61.3\% \\ \hline
VGG-19 & \textbf{66.9\%} & 65.4\% & 64.4\% & 64.2\% & 63.4\% & 64.3\% & 62.9\% & 65.3\% & 61.1\% \\ \hline
\multicolumn{9}{c}{Caltech-256} \\ \hline
VGG-11 & \textbf{29.4\%} & \textbf{29.4\%} & 28.8\% & 29.2\% & 28.3\% & 28.2\% & 28.6\% & 28.9\% & 27.8\% \\ \hline
VGG-13 & \textbf{29.8\%} & 29.5\% & 28.0\% & 29.1\% & 28.9\% & 29.4\% & 29.1\% & 29.1\% & 28.1\% \\ \hline
VGG-16 & 30.3\% & \textbf{30.5\%} & 28.1\% & 27.5\% & 29.4\% & 29.2\% & 30.3\% & 29.6\% & 28.4\% \\ \hline
VGG-19 & \textbf{30.8\%} & 30.7\% & 28.5\% & 28.7\% & 29.0\% & 29.9\% & 28.9\% & 29.2\% & 28.3\%
\\ \hhline{==========}
\end{tabular}}
\end{center}
\end{table*}

\subsection{Performance of ALR on deep networks}
Table \ref{tab:tikhonovcomparison} compares ALR results with other regularization schemes based on the different Tikhonov regularization schemes proposed for deep and shallow neural networks in the recent literature. As one can see, ALR has the best accuracy, and in some cases, the difference between the ALR's accuracy and that of the others is $4\%$.
\section{Conclusion}
This paper studied the effects of adaptive LRF algorithms for neural network regularization entitled DLR and ALR. DLR can control the overfitting in shallow neural networks. Although DLR can find a suitable set of trained weights for a deep neural network, its performance is not perfect. Thus, we extended DLR to ALR. ALR combines DLR and Tikhonov-based regularization approaches. Both DLR and ALR do not affect all network weights when overfitting occurs. They compute condition numbers for the layers of the learning model. By a probability distribution related to condition numbers, some matrices are substituted with their low-rank factorization explicitly by DLR and implicitly by ALR. Selecting a subset of weights for training improves the training convergence and significantly improves the generalization. The most critical point in this adaptive regularization is lazy weights. They do not contribute to the regularization until the last training epochs, so they have no occasion for tuning-up. We use a damping sequence to increase the weights selection probability when the number of epochs increases. This contribution improved ALR significantly because the different layers contribute to regularization by significant likelihood. We tested DLR and ALR on the various networks and datasets and showed that DLR controls the overfitting in shallow networks, and ALR does the same for deep networks.

\section*{References}

\bibliography{ref}

\appendix

\section{Proof of Theorem 1}\label{app:th1}
The goal of training is to minimize the distance between the outputs of the network $f(x;W) =[f(x_i;W)]_{i=1,...,m}$ and the targets by minimizing the following mathematical expectation:

\begin{equation*}
\min_W \mathbb{E}(x;W)=\mathbb{E}[\|f(x;W) - f(x;W^*)\|_2^2]
\end{equation*}

It is direct:
\begin{equation*}
\mathbb{E}[\|f(x;W) - f(x;W^*)\|_F^2]=\mathbb{E}[\|(W - W^*)x\|_2^2]
\end{equation*}

For any $a,b$ and Leibniz integral rule \cite{flanders1973differentiation}, we have:
\begin{equation}\label{eq:th-k-1}
\frac{\partial \mathbb{E}(x,W)}{\partial W_{a,b}}=2 \mathbb{E}[\sum_j(W(a,j) - W(a,j)^*) x(j) x(b)]
\end{equation}
Then:
\begin{equation}\label{eq:th-k-10}
\nabla_{W(a,b)} \mathbb{E}(x,W)= 2 \sum_j(W(a,j) - W(a,j)^*) \mathbb{E}[x(j) x(b)]
\end{equation}

Now we have to compute $\mathbb{E}[x(j)x(b)]$ in two cases:
\begin{enumerate}
\item $j = b$:
\begin{equation*}
E[x(b)^2] = E[x(b)^2] - E^2[x(b)] = \sigma^2
\end{equation*}
\item  $j \neq b$: Since the $x(j)$ and $x(b)$ are independent, then:
\begin{equation*}
\mathbb{E}[x(j)x(b)] = \mathbb{E}[x(j)]\mathbb{E}[x(b)] = 0
\end{equation*}
\end{enumerate}

Therefore, Eq.\ref{eq:th-k-10} can be summarized as following:

\begin{equation}
\nabla_{W(a,b)} \mathbb{E}(x,W)= 2 (W(a,b) - W(a,b)^*)\sigma^2
\end{equation}

In optimal condition of LRF, we have:

\begin{equation}\label{eq:th-k-2}
\nabla_{W(a,b)} \mathbb{E}(x,W) = 2 \sigma^2 (U\Sigma V^T)(a,b) - (U_k\Sigma_k V^T_k)(a,b)]
\end{equation}

We know that $W^* = \sum_{i=1}^n u_i^T \sigma_i v_i$ and $W_k = \sum_{i=1}^{k} u_i^T \sigma_i v_i$ then $W^* - W_k = U\Sigma V^T - U_k \Sigma_k V_k^T = \sum_{i=k+1}^n u_i^T \sigma_i v_i$. Therefore, from Equ.\ref{eq:th-k-2}, we have:

\begin{equation}\label{eq:th-k-5}
\nabla_{W(a,b)} \mathbb{E}(x,W) = 2 \sigma^2 \Big(\sum_{i=k+1}^n (u_i^T \sigma_i v_i)\Big)(a,b)
\end{equation}

By using gradient descent rule to update $W_k$, we have the following where step-size is $\frac{1}{2\sigma^2}$:

\begin{equation*}
W_k = W_k + \gamma \nabla_W \mathbb{E}_x [\| f(x,W) - f(x,W^*)\|_F^2]
\end{equation*}

Therefore, we have:

\begin{equation*}
W_k = W_k + \gamma (2\sigma^2 (\sum_{i=k+1}^n u_i^T \sigma_i v_i)) = W_k + \sum_{i=k+1}^n u_i^T \sigma_i v_i = W^*
\end{equation*}

And the proof is completed.
\qed
\section{Proof of Lemma 1}\label{app:lem1}
This lemma is proved by contradiction. Assume that the rank of $\tilde{W}$ is one. Because of linear dependency between the rows, there is an $i\in\{1,...,m\}$, such that:
\begin{equation}\label{eq:lemma1eq1} \tilde{W}^i = \sum_{j \neq i}\alpha_{i,j} \tilde{W}^j. \end{equation}
where $\tilde{W}^l$ shows the $l^{th}$ row of $\tilde{W}$. 

Consider the samples of $i^{th}$ class, such as $(x_k, y_k)$. It meets $y_k^i = 1$ and for the others $y_k^j = 0$. To satisfy the linear model, we have:
\[\tilde{W}x_k = y_k\iff \tilde{W}^j x_k = y_k^j, \forall j=1,...,m.\] 
Thus $\tilde{W}^i x_k = 1$ and for $\forall j \neq i, \tilde{W}^j x_k = 0$. Based on Eq. (\ref{eq:lemma1eq1}), we have:
\begin{equation*}
1 = \tilde{W}^i x_k=\Big(\sum_{j\neq i}\alpha_{i,j}\tilde{W}^j \Big)x_k = \sum_{j\neq i}\alpha_{i,j}(\tilde{W}^j x_k) = 0,
\end{equation*}
This contradiction proves the theorem and $Rank(\tilde{W}) > 1$.
\qed

\section{Proof of Theorem 3}\label{app:th3}
Consider the loss function with the LRF regularization term:
\begin{equation*}
E^*(\theta_t) = E(\theta_t) + \gamma \|w^* - LRF(w^*)\|_F^2
\end{equation*}
where $\theta_t$ is a trained weight and $w^*$ is a lazy weight. The gradient of this function with respect to $\theta$ after appearing the lazy weight is:
\begin{equation*}
\nabla_\theta E^*(\theta_t) = \nabla_\theta E(\theta_t) + \nabla_w \gamma \|w^* - LRF(w^*)\|_F^2
\end{equation*}
Since the network before the lazy weight is trained, we have $\nabla_\theta E(\theta_t) \approx 0$, then:

\begin{equation*}
\nabla_\theta E^*(\theta_t) =2\gamma \|\sum_{i=1}^r u_i^T\sigma_i v_i - u_r^T\sigma_r v_r\|_F,
\end{equation*}

where $\sum_{i=1}^r u_i^T\sigma_i v_i$ is the singular decomposition of $w^*$, $\sigma_i$ is its singular values and $u_i$ and $v_i$ are singular vectors. Without losing of the generality, we assume that the rank of $w^*$ is $r$. Also $LRF(w^*)$ is defined by 1-rank factorization $u_r^T\sigma_r v_r$. It is obvious that:

\begin{equation*}
\nabla_\theta E^*(\theta_t) = 2\gamma \sqrt{\sum_{i=1}^{r-1} \sigma_i},
\end{equation*}

Because $w^*$ is a lazy weight and LRF is not imposed on this weight, based on Lemma 1, its rank is greater than 1, thus there is at least $\sigma_{i\in\{1,...,r-1\}} > 0$. Then, at the extremum point, the value of the gradient is great and it drops the accuracy atleast by $\sigma_{i\in\{1,...,r-1\}}$.
\qed

\end{document}